\documentclass{article}
\usepackage{graphicx} % Required for inserting images
\usepackage{booktabs}
\usepackage{longtable}
\usepackage{array}
\usepackage{listings}
\usepackage{xcolor}
\usepackage{svg}
\usepackage{authblk}
\usepackage[margin=1.5in]{geometry}

\title{Research Assistant: AstraZeneca's Agentic System for R\&D}
\author[1]{Piotr Grabowski\thanks{Corresponding author: piotr.grabowski1@astrazeneca.com}}
\author[1]{Mohamed Alameen}
\author[1]{Jorge Bretones}
\author[1]{Sabina Cardell}
\author[1]{Miguel Carmona}
\author[1]{Gavin Edwards}
\author[1]{Ben Grainger}
\author[1]{Sameh Hassan}
\author[1]{Erik Jansson}
\author[1]{Artur Kuziakhmetov}
\author[1]{Albert Maristany}
\author[1]{Hebatallah Mohamed}
\author[1]{Andriy Nikolov}
\author[1]{Sebastian Nilsson}
\author[1]{Mark O'Donoghue}
\author[1]{James Pacileo}
\author[1]{Ashiq Sultan}
\author[1]{Alex Voegele}
\author[1]{Michaël Ughetto}

\affil[1]{AstraZeneca, RDIT, Biological Insights Knowledge Graph}

\date{}

\begin{document}

\maketitle

\begin{abstract}
We describe Research Assistant, an internal LLM-based system developed at AstraZeneca to help scientists and clinicians explore biomedical questions across a broad range of data sources. The system provides a chat-style interface that brings together evidence from scientific literature, knowledge graphs, chemistry, clinical trials, safety resources, expression data, and internal experimental systems. It supports both a fast mode for direct question answering and a multi-step mode for more complex research tasks. Responses are grounded in retrieved evidence and linked back to the original sources, allowing users to review and further explore the underlying data. In this technical note, we outline the system architecture, the main design choices behind the product, and lessons learned from deploying it at scale to support day-to-day R\&D workflows across AstraZeneca.
\end{abstract}

\section{Introduction}

With the introduction of ChatGPT by OpenAI in the second half of 2022, a new way of interacting with information systems was unlocked. Framing this interaction as a series of natural language queries and responses made these systems more accessible to non-technical users. Prior to the release of ChatGPT, within our team at AstraZeneca we built and maintained an array of data services for consuming our Knowledge Graph \cite{Geleta2021.10.28.466262} and NLP pipeline to support multiple drug discovery programmes. However, these resources require a certain level of software engineering experience to use, for example by writing queries for the REST APIs or querying the graph using the Cypher query language. The introduction of ChatGPT prompted us to reconsider how these resources could be accessed by scientific and clinical experts at AstraZeneca, with the aim of reducing technical barriers. We developed Research Assistant to make internal biomedical data resources more accessible to domain experts. It provides a chat-style interface for biomedical questions and returns citation-linked answers grounded in multiple data sources. Research Assistant grew from a small pilot to 15,000 unique internal users within one year.  Here we describe the system and share lessons learned along the way.

Research Assistant is an LLM-based multi-agent system for the biomedical domain with the following workflow:
\begin{enumerate} 
    \item Given a user question, find the most appropriate data sources and facts.
    \item Ground the answer in retrieved evidence.
    \item Link users to the original sources for manual evaluation and further exploration.
\end{enumerate}

\section{Architecture}
\subsection{Application - High-Level Overview}
\begin{figure}
    \centering
    \includegraphics[width=0.85\linewidth]{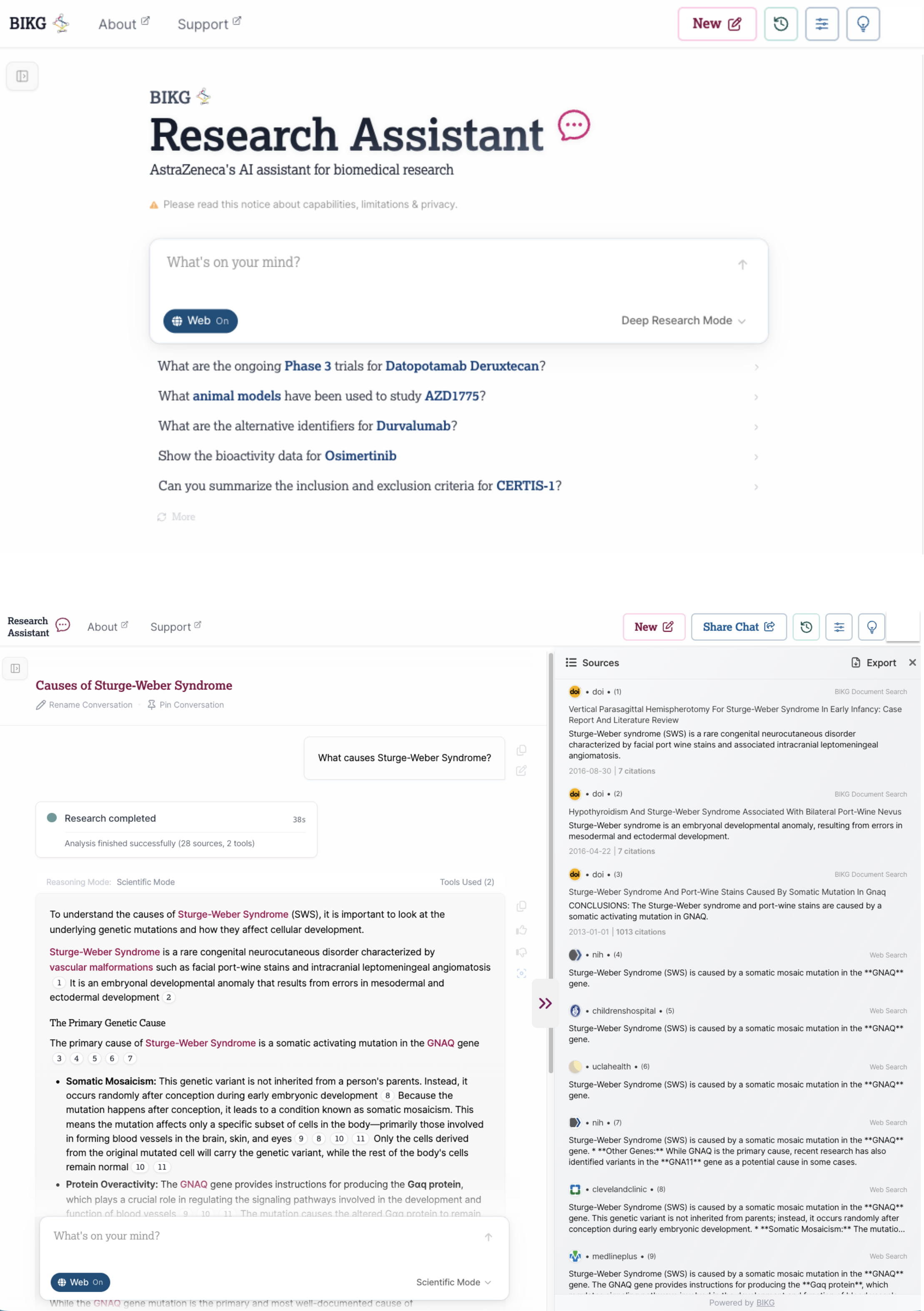}
    \caption{Screenshot of the Research Assistant User Interface (UI). The upper panel shows the initial screen where the user can submit the query. The lower panel shows an example answer using the Scientific Mode to a question "What causes Sturge-Weber Syndrome?. The right open panel contains citations used by the system to generate the answer. The automatically highlighted entities in the text response link to additional resources on genes, diseases and chemistry."}
    \label{ui_screenshot}
\end{figure}

The backend for Research Assistant was written in Python with heavy use of the asyncio package. The layer serving the API endpoints was developed in FastAPI. The framework used at the heart of the application was Apache Burr (https://burr.apache.org/). Apache Burr is a light-weight state machine framework which allows to build complete assistant-style applications involving loops, conditional execution of parts of the application graph and offers out-of-the-box telemetry support without binding to any specific vendors. Applications in Apache Burr are defined as graphs in which each node is an Action and each edge is a transition between Actions. An Action is essentially a Python function where a modification of the application's State is performed. The frontend was written in TypeScript and React. During the state machine execution the backend streams messages with state updates (using server-sent events) asynchronously to the frontend, updating the various UI elements in real time and informing the user about the status of their query. Example screenshots are shown in Figure \ref{ui_screenshot}.

In Research Assistant, the primary contributors to system performance are the data services. We believe the quality of the underlying data services and APIs contributed substantially to internal adoption. Keeping the LLM instructions lean while guiding the system to rely only on the obtained external information was intended to reduce hallucinations, which are more common in systems without grounding data. The system uses a parallel architecture in which a user query triggers multiple lightweight tool calls simultaneously. A final synthesis step is then performed by a larger LLM. This approach also helped keep per-query costs low. With current per-token pricing by Google Cloud Platform (July 2026) the average cost of a Research Assistant query is only 16 cents (including all input and output tokens, using Gemini 3 Flash and Gemini 3.1 Pro models). Another advantage of this retrieval-heavy, token-efficient design is lower latency. While it cannot compete with vanilla LLM queries, responses are typically generated within 10 to 30 seconds, which allows Research Assistant to be both an interactive system and work as a source of information for other applications via its REST API and the Model Context Protocol (MCP) endpoints.

\begin{figure}
    \centering
    \includegraphics[width=1\linewidth]{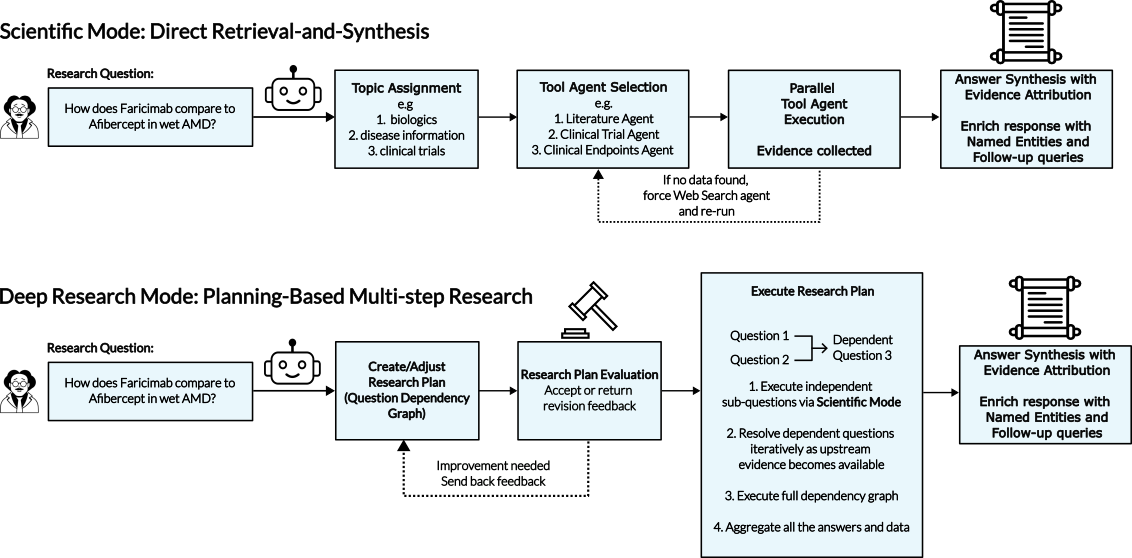}
    \caption{Simplified application graphs of the two main modes of Research Assistant. The Scientific Mode is a single-pass retrieval and synthesis workflow for retrieving information on various biomedical topics. The Deep Research Mode is more complex and used for questions that require multiple rounds of Scientific Mode workflow orchestrated by the research plan executor agent.}
    \label{modes_figure}
\end{figure}

\subsection{Application Modes}
\begin{figure}
    \centering
    \includegraphics[width=1\linewidth]{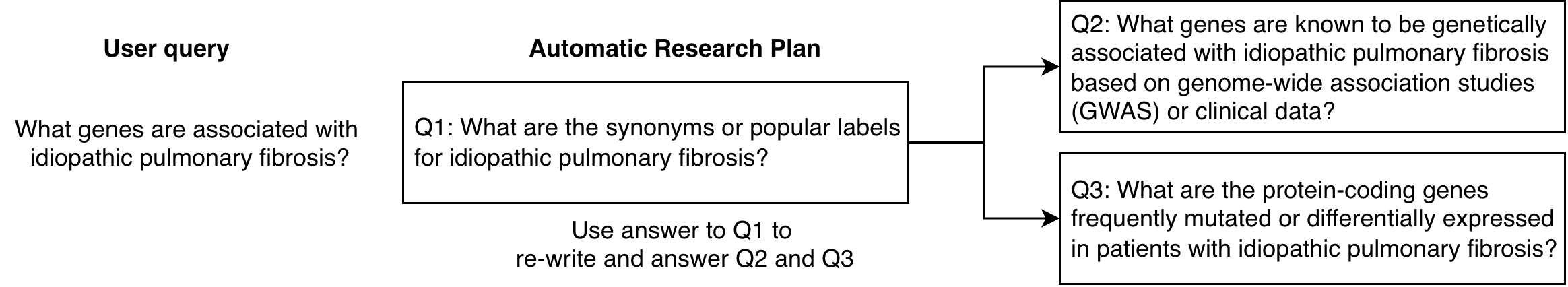}
    \caption{Example of a simple automatic research plan created for the user query "What genes are associated with idiopathic pulmonary fibrosis?"}
    \label{plan_example}
\end{figure}

The entry point for every interaction with Research Assistant is a user question. Two modes are available to users: "Scientific Mode" and "Deep Research Mode" (Figure \ref{modes_figure}). The Scientific Mode balances the complexity and depth of the answer with speed by performing a single feed-forward pass mapping the user query to relevant tool agents (each of which is specialised in different topics and data) and answers the query based on the discovered data. 
In contrast, the Deep Research Mode offers a multi-step planning and execution workflow. It's built around the concept of a research plan which is modeled as a directed acyclic graph (DAG) of questions (Figure \ref{plan_example}), simulating how a real human would approach a more complex problem.

The application initially creates and revises the plan until the judge agent accepts it (with a maximal assigned number of revisions to avoid very long planning loops). The accepted DAG of questions is then topologically sorted such that the independent, initial questions can be answered first, and then dependent questions will be asked and answered as soon as the previous questions were processed. Each of the questions processed in this research DAG is itself a question asked by the application to itself, but using the Scientific Mode. During execution of the research DAG, dependent questions are rewritten using information obtained in earlier steps. For example, a later query about a drug may use a synonym identified in a previous answer. Furthermore, this formulation of the deep research-type system allowed it some flexibility. Users can even ask questions about repeating some analysis on multiple different entities like genes, drugs or clinical trials and the flexible planner will translate this request into an iterative research plan. One can even provide the application with a very specific research plan and flow of questions to override the automatic plan generation step. Importantly, this approach gives more control over the complexity and time of the execution as the size of the research plan is fixed at the beginning. This can be important for research runs which have to finish within a specific amount of time and budget.

\subsection{Tool Agents}
The Tool Agents (listed in Table 1) are central to Research Assistant's ability to answer diverse sets of questions and discover relevant data. A Tool Agent is composed of two things: the tool API and the prompt (similar to the concept of Skills). The tool API is a simple interface which translates things like keywords, accession numbers, IDs and various filter settings into function calls accessing either internal AstraZeneca resources or external APIs. The role of the prompt is to explain to the LLM how to best map the free-text user query into a structured tool API query, what are the limitations of each tool and how the returned data is formatted. 
The glue between the Tool Agent LLM calls and the many tool APIs is the Instructor package (https://github.com/567-labs/instructor/). Instructor facilitates the creation and validation of the structured Pydantic models by the LLM and handles the retries in case of validation errors. Using the Instructor package, together with the recent improvements in structured output generation by LLMs (e.g. JSON format), meant that the queries rarely fail due to model schema validation errors or malformed JSON objects.

Each agent returns the data packaged into the same Pydantic class instance (Observation) containing:
\begin{itemize}
    \item ID of the observation (a tag used later to inject into the LLM response for grounding each statement created by the LLM)
    \item URL for the source of the data, allowing the users to inspect the sources manually
    \item flexible container with grounding data containing sentences, tabular data, knowledge graph triples, etc.
    \item citation string used by the front-end to display it to the user
\end{itemize}

Example hit from the Literature Agent for query \textbf{"Is NRF2 pathway linked to inflammatory bowel disease?"}:

\begin{lstlisting}
Observation(
    id='lit-5',
    link='https://doi.org/10.1186/s12935-022-02660-5',
    data={
        'excerpt': [
            'In inflammatory bowel disease, Nrf2/HO-1 expression decreased MMP-9 and MMP-7, finally helping to reduce inflammation; The skin damage caused by UV irradiation is more severe in Nrf2 knockout mice than in WT mice, and MMP-9 and MIP-2 (macrophage inflammatory protein-2 is a crucial modulator of neutrophil recruitment) levels are much higher [150, 151].'
        ],
        'document_title': 'The Molecular Biology And Therapeutic Potential Of Nrf2 In Leukemia',
        'document_source': 'Cancer Cell International',
        'date': '2022-07-29',
        'document_authors': [
            'Atefeh Khodakarami',
            ...
            'Farhad Jadidi-Niaragh'
        ],
        'number_of_citations': 27,
        'impact_factor': 6.91
    },
    citation='Atefeh Khodakarami et al. - 2022-07-29 - The Molecular Biology And Therapeutic Potential Of Nrf2 In Leukemia - Published in: Cancer Cell International - Citations: 27, Impact Factor: 6.91'
)
\end{lstlisting}

\setlength{\heavyrulewidth}{0.08em}
\setlength{\lightrulewidth}{0.05em}

\begin{longtable}{p{0.20\textwidth} p{0.38\textwidth} p{0.38\textwidth}}
\caption{Summary of Tool Agents. Topics are used by the tool selection routine which discovers topics within user queries. Presence of any of these topics in the query will lead to adding that specific Tool Agent to the list of agents run for the query.} \label{tab:tool-agents} \\
\toprule
\textbf{Agent Name} & \textbf{Use-Case} & \textbf{Topics} \\
\midrule
\endfirsthead

\multicolumn{3}{c}{\tablename\ \thetable{} -- continued} \\
\toprule
\textbf{Agent Name} & \textbf{Use-Case} & \textbf{Topics} \\
\midrule
\endhead

\midrule
\multicolumn{3}{r}{Continued on next page\ldots} \\
\endfoot

\bottomrule
\endlastfoot

Literature Agent & Scientific literature search, electronic notebooks, patent/conference lookups. & biological relationships, chemistry information, drug safety, general biomedical knowledge, internal experimental data \\[4pt]
\midrule

Compound Agent & Compound ID lookups, SMILES resolution, bioactivity, physicochemical properties. & chemistry information, entity synonyms \\[4pt]
\midrule

Knowledge Graph Agent & Pairwise entity relationships (gene--disease, compound--target, compound--disease). & biological relationships \\[4pt]
\midrule

Clinical Trial Agent & Trial search by drug/condition/status/sponsor; specific NCT lookups, inclusion/exclusion criteria. & clinical endpoints, clinical trials \\[4pt]
\midrule

Web Search Agent & Real-time web search; recent organised events and news. & general biomedical knowledge, recent events \\[4pt]
\midrule

OFF-X Agent & Drug adverse events and safety alerts by drug name or gene target. & drug safety \\[4pt]
\midrule

Discover Agent & Ranked predictions of novel gene--disease--compound associations. & discovery and ranking \\[4pt]
\midrule

Mapping Agent & Cross-database ID and synonym mapping for compounds, genes, diseases. & entity synonyms \\[4pt]
\midrule

Clinical Endpoints Agent & Clinical efficacy endpoint extraction (OS, PFS, ORR, CR) with LLM-generated summary. & clinical endpoints \\[4pt]
\midrule

Human Protein Atlas Agent & Tissue and cell-type gene expression levels (broad patterns or specific TPM/CPM values). & gene expression \\[4pt]
\midrule

Glossary Agent & AstraZeneca-specific acronym and abbreviation definitions. & AZ terminology \\[4pt]
\midrule

In Vivo Agent & Internal AZ in vivo study data: toxicity studies, animal models, compound testing, dosing. & internal experimental data, preclinical studies \\

\end{longtable}

\subsubsection{Literature Agent}
The Literature Agent is the most important Tool Agent in Research Assistant. It connects the system with an internal NLP pipeline created and maintained within the same team. The NLP pipeline relies on CoreNLP \cite{coreNLP} from Stanford NLP (https://github.com/stanfordnlp). Briefly, the NLP pipeline ingests over 3.8 billion sentences from over 67 million documents covering sources such as PubMed, EuropePMC, Embase, bioRxiv, medRxiv, Wiley, Springer Nature, Dailymed, Insightmeme and AstraZeneca's internal Electronic Lab Notebook (ELN) systems. 
The pipeline consists of three main steps:
\begin{enumerate} 
     \item Deep-learning based biomedical named entity recognition (NER) using the KAZU pipeline \cite{kazu}
     \item Deep-learning based grammatical and syntactic structure extraction
     \item Rule-based relation extraction (RelEx)
\end{enumerate}
The processed sentences are then indexed for fast retrieval using the Elasticsearch engine. To account for spelling and naming variation across entities we employed a two-fold approach. First, we index the lemmatized versions of words contained in the sentences. In this way, words such as ‘gene’ and ‘genes’ in different sentences would be represented as a lemmatized form "gene". When the Tool Agent creates sets of keywords to use for searching, it uses the lemmatized versions of query keywords. Secondly, since the sentences in the NLP index already have annotated Named Entities, such as Ensembl gene IDs or MONDO IDs for diseases, we employ an automatic translation of the Tool Agent-created keywords into Named Entities using the KAZU pipeline \cite{kazu} and add them to the set of keywords created by the Agent. This allows us to increase recall when retrieving sentences from the Elasticsearch index. The LLM creates two types of keywords: "must" keywords and "should" keywords. The "must" keywords are all required to be in one sentence to consider it a hit, whereas the "should" keywords don't have to be present, but do increase the score of that sentence if they are. Our approach offers automatic retrieval of specific statements found in the biomedical literature. As the literature pipeline indexes and retrieves individual sentences, this gives the model a more focused evidence set than abstract-level retrieval and may help the LLM to focus on the most relevant parts of the context without filling it with less relevant content while consuming substantially smaller numbers of input tokens. For publication searches, after filtering the sentences on LLM-generated keywords, a preliminary set of hits is enriched with information from OpenAlex \cite{openalex}. Specifically, the "citation normalized percentile" score (publication citation count normalized by work type, year, and subfield) and the percentile-transformed journal-level h-index are appended. These OpenAlex metrics are then summed together with the percentile-transformed Elasticsearch scores to create one final relevance score used for sorting the sentences. This heuristic is used to combine the sentence semantic relevance and publication impact to avoid returning sentences which might contain required "must" keywords, but come from low-impact journals or very low-cited publications.

\subsubsection{Compound Agent}
The Compound Agent is a linker between the Research Assistant and AstraZeneca's Chemistry Application Gateway (CAG). CAG is a data source serving information about structures, bioactivities, synonyms, patents and much more. This Tool Agent communicates with the service via a set of templated GraphQL queries which the LLM is tasked to complete based on the user query. It fetches information on activity values from biochemical assays, compounds' physicochemical properties, translates synonyms of compounds, and translates SMILES into specific compound names, giving Research Assistant the ability to contextualize its answers with data from both internal and external chemistry sources. 

\subsubsection{Knowledge Graph Agent}
The Knowledge Graph Agent gives Research Assistant access to our Biological Insights Knowledge Graph \cite{Geleta2021.10.28.466262}. GraphRAG has become a popular method in recent years to ensure question answers are grounded using reliable factual data and reduce hallucinations \cite{graphrag}. In this approach, an LLM agent having the knowledge of the graph schema receives the user's natural language question like ``What genes are important for atopic dermatitis?'' and translates it into a structured graph query like 

\begin{verbatim}
    MATCH (g:GeneTarget)-[:ASSOCIATES]->(d:Disease)
        WHERE d.default_id = "MONDO:0004980" 
        RETURN g.default_id, g.default_label
\end{verbatim}

However, we found that a straightforward approach, which involves letting an LLM access the whole graph schema and directly generate Cypher/SPARQL graph queries, leads to several issues when dealing with a sufficiently large and complex integrated graph like BIKG:
\begin{itemize}
    \item {Behaviour becomes non-deterministic: the LLM can interpret the same question in slightly different ways and explore different paths in the graph trying to answer it, which can in turn lead to different ranges of answers. For instance, "genes associated with a disease" in one case can include direct experimentally supported Gene Target $\rightarrow$ Disease edges
    
    \begin{verbatim}
        MATCH (g:GeneTarget)-[:ASSOCIATES]->(d:Disease) ...
    \end{verbatim}
    but in another case can also include inferred paths via intermediate pathways or drugs in trial
    
    \begin{verbatim}
        MATCH (g:GeneTarget)-[:PARTICIPATES]->(p:Pathway)
            -[:INVOLVED_IN]->(d:Disease) ...
    \end{verbatim}
    
    }
    \item {Difficulty taking into account meta-level information. Different data sources and even different facts in the graph can have different levels of confidence: e.g., one gene-disease association edge may reflect a genuine causal impact while another one only a change in expression level. While including low confidence information can be justified in an analytical workflow, where the confidence factors can be taken into account and adjusted for, it can mislead the user in a question answering context. Selection of appropriate data sources based on trust in a consistent way also presents a difficulty for an LLM.}
    \item {Generated queries can be factually correct, but too complex and lead to timeouts. It is difficult for an LLM to build high-performance queries based only on the schema.}
\end{itemize}

For this reason, when implementing our Knowledge Graph Agent, we adopted several techniques to streamline the process and provide guardrails to improve its reliability and consistency.
The Knowledge Graph Agent's workflow involves the following steps:
\begin{enumerate}
    \item{Grounding the entities mentioned in the question. Named entity mentions (e.g., gene names, diseases, tissues) get resolved to their IDs using the KAZU NER service and the BIKG mapping service (an internal microservice for translating ID systems for entities using data derived from the graph build process).
    At this stage, the original question like ``What genes are important for atopic dermatitis?'' gets tagged with the recognized entity ID (``atopic dermatitis'' = ``MONDO:0004980'').
    }
    \item{Creating a prototype query based on a projected schema. The projected BIKG schema describes a subset of the BIKG graph targeted for the question answering use case, including the most relevant node types, relations, and attributes. The prototype query defines the matching pattern over the graph in terms of this projected schema. At this stage, the system constructs the matching clauses of the query like 
    
    \begin{verbatim}
        MATCH (g:GeneTarget)-[r:ASSOCIATES]->(d:Disease) ...
    \end{verbatim}
    
    }
    \item{Adjusting the prototype query using deterministic rules. These rules include, for instance, relevant sources to be picked for each edge type, categorical attributes describing specific fine-grained concepts (like "phosphorylation", "inhibition", etc.), or descriptive node attributes to be returned for every query. In our case, it can restrict the data sources to the most relevant ones

    \begin{verbatim}
        MATCH (g:GeneTarget)-[r:ASSOCIATES]->(d:Disease) ...
        WHERE 
            (d.default_id = 'MONDO:0004980')
            AND (r.prov IN ['OMIM_DATA', 'ORPHANET'...])
    \end{verbatim}
    
    }
    \item{Executing the final query and returning a summary of results. After all relevant rules are applied to the prototype query, it gets executed over the BIKG Neo4j endpoint and its results are returned to the caller agent.}
\end{enumerate}

\subsubsection{Clinical Trial Agent}
The Clinical Trial Agent retrieves data from ClinicalTrials.gov via the REST API (https://clinicaltrials.gov/data-api/api). It can either retrieve data on specific trials given the presence of NCT IDs in the user query, or use the filtering capabilities of the ClinicalTrials.gov REST API to generate sets of filters and search by diseases, drugs, sponsor names and other additional terms. Because trial records can be large, we applied a compositional approach. This approach allows for fetching only the specific pieces of trial data that are relevant to the user's query, rather than downloading the entire dataset for each trial. The LLM selects the necessary categories based on the user's request. Each category corresponds to a specific section of the clinical trial data, such as 'Basic Information', 'Eligibility and Participant Criteria', 'Outcomes and Endpoints', or 'Adverse Events and Publications'. By composing the final result from these requested data blocks, the Tool Agent significantly reduces the token size of the retrieved data for each trial.

\subsubsection{Web Search Agent}
The Web Search Agent is used to provide information about recent events and extend Research Assistant with additional external data, especially when web scraping is not available to enterprise teams for legal reasons. This particular implementation uses the Google Vertex AI platform to perform Grounding with Google Search with the Gemini 3 Flash model. The Vertex AI response contains grounding markers showing which parts of the Gemini response map to which web sources. This allows the agent to turn the Gemini response into a set of Observation objects, described earlier, and be treated by Research Assistant like any other tool. While it is useful for surfacing information on most recent events and often on high-quality scientific literature, in practice we notice that the Web Search Agent can at times access sources that are not scientifically credible, including discussion forums and various blog posts. Luckily, this can be controlled with a domain exclusion list. Moreover, even with thinking levels set to "medium", the Google search Gemini responses can claim facts as grounded in specific web resources, while a closer inspection of those resources might show them only tangentially mentioning a related fact. Additional care and verification of sources is required when dealing with responses grounded mostly in such web search data.

\subsubsection{Discover Agent}
While Research Assistant is primarily designed to retrieve information about known relationships and established scientific facts, we extended the system to support predictive capabilities. To this end, we developed a platform for link prediction in the Biological Insights Knowledge Graph using graph machine learning approaches, including:
\begin{itemize}
    \item Graph Convolutional Neural Networks \cite{gat} \cite{rgcn} 
    \item Knowledge graph embedding models (RotatE \cite{rotate} , ComplEx \cite{complex}) 
    \item Path-based methods (Degree-Weighted Path Count \cite{dwpc}, S\"orensen Similarity)
    \item PageRank
\end{itemize}

Each model generates a ranked list of candidate links which can suggest how likely is it that a given gene is associated with some disease, for example. These rankings are then combined using CRank \cite{crank} to identify predictions with the strongest agreement across models and returned to Research Assistant. This Tool Agent is particularly useful in Deep Research Mode. In this setting, the system first generates a prediction using the Discover Agent and then retrieves evidence supporting or contradicting that relationship. Because Deep Research Mode follows a predefined execution plan, this workflow remains bounded in both runtime and token budget.

\subsubsection{Clinical Endpoints Agent}
The agent for clinical endpoints is based on a ClickHouse database produced by the NLP Endpoints pipeline. This pipeline takes the normalised NLP-parsed data for the following sources:
PubMed, Europe PMC, medRxiv, bioRxiv, Wiley, Springer, and InsightMeme and applies four models to them, based on Google's BERT model \cite{bert}. The four models were trained to extract entities from scientific literature relating to drug properties and clinical trials read-outs, such as RECIST, drug efficacy, drug safety, and pharmacokinetic and pharmacodynamic drug properties.
The models were trained and tested on plain text from Dailymed (https://dailymed.nlm.nih.gov/dailymed/), Pubmed and Europe PMC. The extracted sentences were annotated by either LLM or regular expression (regex) rules, and quality controlled by manual inspection. F1 score on the test set was used to evaluate the fine-tuned models, and once a model reached F1 of at least 0.85 it was deemed sufficient for deployment.

An example of a sentence with efficacy endpoints labelled:
 
\definecolor{oscolor}{RGB}{255,230,153}
\definecolor{csscolor}{RGB}{189,215,238}

\newcommand{\oshl}[1]{\colorbox{oscolor}{#1}}
\newcommand{\csshl}[1]{\colorbox{csscolor}{#1}}

The five year \oshl{overall survival (OS)} rate was \oshl{59.3\%} for surgery plus chemotherapy and \oshl{65.9\%} for surgery only, whereas the \csshl{cancer-specific survival (CSS)} rates were \csshl{61.8\%} and \csshl{73.5\%}, respectively.

\subsubsection{Mapping Agent}
The Mapping Tool Agent extracts various entities from user queries and retrieves synonyms associated with these entities from the BIKG node index. For example, it allows Research Assistant to recognize that EGFR is also called ERBB or HER1. This is crucial when the system is preparing a broader search using the Literature Agent but the user query only contains one of the synonyms. Construction of the BIKG graph~\cite{Geleta2021.10.28.466262} also includes resolving various available identity mappings between concepts from different ontologies, as well as bringing together known synonyms for the same concepts: e.g., different names for a disease, alternative identifiers for a drug (research ID vs. trade name). This involves selecting a list of ``canonical'' entities, deciding whether two entities are identical and can be merged, and aggregating various alternative names. The output of this process is an index of biological entities with their canonical IDs and names, as well as alternative identifiers and synonyms. The Mapping Agent utilises this index to resolve mentions of biological entities in users' questions and ground them to the corresponding concepts in the graph. This grounding can be used both directly (e.g., for simple questions like "What is the compound name for Tagrisso?") as well as indirectly by other agents: e.g., to enrich Literature Agent requests with synonyms or to find a canonical entity ID to reference in Knowledge Graph Agent Neo4j queries.

\subsubsection{Glossary Agent}
The Glossary Tool Agent is used for translating the various terms used within AstraZeneca. For example, "A\&R" in the user query might mean "Analysis and Reporting" or "Alliance and Operations". This can help the LLM agents navigate the complex acronym systems used within the pharmaceutical and corporate environment. This is a very simple agent whose entire data backend is a single SQL table and can be easily expanded whenever a new acronym or term is being introduced.

\subsubsection{Human Protein Atlas Agent}
The Human Protein Atlas (HPA) Tool Agent is tasked with obtaining expression patterns of genes from an established public resource by connecting to the HPA REST API and linking back the users to this resource. Gene and protein expression information is crucial for drug discovery R\&D and allows the system to contextualize its responses, e.g., when interpreting link predictions between a gene and disease from the Discover Agent. Access to this information is especially useful for the multi-step Deep Research Mode where prompts may require retrieval to be limited to genes expressed in a specific cell type or organ, for example "What classes of compounds affect genes expressed in the liver?". 

\subsubsection{In Vivo Agent}
The In Vivo Agent was added to Research Assistant as a way for AstraZeneca scientists to discover preclinical data from drug experiments and, as such, has a narrower scope compared to the other described agents. While the main focus of our system is on factual data, this integration is part of the efforts to make all types of experimental data (including omics, imaging, etc.) findable and reusable, even when the user was not specifically searching for them. This gives the existing datasets a higher chance of being re-used which leads to reduced operational costs and need for animal studies.

\subsubsection{OFF-X Agent}
The role of the OFF-X Tool Agent is to retrieve information on drug safety alerts and other safety-related information from the Clarivate OFF-X API, including safety profiles of specific gene targets. This information is predominantly used to enrich the data from other tool agents with most recent high-quality safety data.

\subsection{Tool Agent Picker}
The Tool Agent Picker is a router module responsible for mapping the user query to sets of Tool Agents. This is a crucial step as running all the Tool Agents concurrently for each user query would lead to slower response times due to LLM endpoint limitations. Furthermore, limiting the amount of data fed into the context for each system response improves the quality of the Assistant's answers and decreases the cost of each query.  During early development of the system we experimented with various approaches, as there was no established way to do this. For example, one approach to tool routing used a hand-crafted database of example queries the system was expected to handle well. These example questions were embedded during application build time using OpenAI's text-embedding-3-small model and stored in a ChromaDB database. These exemplar questions would be mapped to the "best suited" Tool Agents. The actual user query would then be embedded at runtime using the same model and the set of Tool Agents linked to the exemplar question with the lowest cosine distance to the user query in the embedding space would be selected. This approach was fast and allowed a potentially unlimited number of exemplars, providing substantial flexibility in tool selection. However, as the system was being used by more diverse users we noticed that our hand-crafted database of exemplar questions became hard to maintain. Furthermore, the nearest neighbour search in the embedding space between the user query and the exemplar questions will always yield a hit, even if the query is not semantically related to the entries in the database, unless one specifies a distance cut-off above which one considers the query an outlier. Defining that cosine distance cut-off given a live and dynamically developed system was very hard and this approach was dropped.

We subsequently adopted a different approach that now supports thousands of monthly interactions. We performed automated topic modeling using machine learning methods on a database containing tens of thousands of real user interactions. This allowed us to shortlist a small number of topics present in typical user queries such as "Chemistry Information", "Biological Relationships", "General Biomedical Knowledge", "Drug Safety" and "Recent Events". We then mapped the best set of Tool Agents which can support answers to these topics. At run time the user query is automatically assigned these predefined topics and the union of the Tool Agents assigned to the selected topics is then run in parallel. This makes the tool selection process more transparent and easier to maintain, while preserving flexibility. Large Language Models are good at high-level topic modeling, but they are not experts in our internal Tool Agents. This added interface of topic-to-agent mapping allows the models to operate on more robust concepts while allowing the developers to decide which Tool Agents are best suited for what kinds of topics.

\section{Evaluations}
    
During the initial phases of development of Research Assistant we used the BioASQ Training 10b question-answer set \cite{bioasq}. Specifically, we used a random subset of 100 yes/no questions which allowed us to calculate balanced accuracy of the system compared to "vanilla" LLMs, like GPT 4 (which was the state-of-the-art model at the start of the project). This allowed us to direct the development of the Research Assistant without any users (the so-called "cold start" problem). However, because such a narrow QA set was not suitable as the primary target metric, we used it mainly as a sanity check and for regression monitoring. The most informative evaluations came from direct user feedback as the Research Assistant was gaining traction with AstraZeneca's employees and the product team could turn user feedback and needs into a feature plan. Importantly, the improvements made in Research Assistant based on user feedback did not translate to performance improvements on the BioASQ question set, further exemplifying the orthogonality of biomedical QA benchmarks and real-life user requirements.

Another evaluation set used throughout the project was the STaRK dataset \cite{wu2024starkbenchmarkingllmretrieval}. Briefly, Wu et al. used a PrimeKG knowledge graph \cite{Chandak_Huang_Zitnik_2023} and simulated many potential user queries about the entities in the graph. This allowed them to construct a collection of natural language queries mapped to sets of PrimeKG nodes. After mapping the entities in PrimeKG to our internal AstraZeneca knowledge graph, the BIKG \cite{Geleta2021.10.28.466262}, we were able to use this benchmark to drive the development of our Knowledge Graph tool agent. Using random subsets of queries from STaRK, we could calculate overlap statistics between the triples returned by the Tool Agent given the input query and the expected node set. This served as a simple benchmark for monitoring the Knowledge Graph tool agent for possible regressions during the development of the Research Assistant.

With the growing internal user base and the database populated with tens of thousands of real user interactions we started using automated approaches to system behaviour analytics. For example, we developed a simple LLM judge agent which, given a pair of real user question and a system response, would assign a score and add comments about that interaction. This allowed us to rapidly identify areas of the Research Assistant where improvements were needed, for example by adding new data sources or tuning the tool agent behaviour.

\section{Programmatic Usage Example - Patient Safety}
An example of programmatic use of Research Assistant at AstraZeneca is the CRAM Auto Tool, which supports Combination Risk Assessment for Patient Safety Scientists. The aim of this process is to evaluate the safety profile when two or more products are used in combination, especially in clinical trials, and to identify new or altered risks compared with the individual products. This requires integrating evidence on target biology, pharmacology, mechanism of action, and clinical findings, with results documented in a structured risk prediction table and updated as new evidence emerges \cite{CRAM}.

A critical step in this workflow is determining whether modulation of a drug target is linked to a biological mechanism that could plausibly contribute to an adverse event. For example, the questions posed to Research Assistant might include:

\begin{itemize}
    \item What pathways are implicated in nausea?
    \item Does the EGFR gene have a role in nausea?
    \item Which biological pathways is EGFR involved in?
    \item Do EGFR inhibitors have a role in nausea?
\end{itemize}

Before this capability was automated in CRAM, using Research Assistant to perform a focused preliminary literature review connecting a target to an adverse event had already become a standard workflow among scientists. The CRAM Auto Tool embedded this established usage pattern into the product by calling Research Assistant programmatically. Research Assistant retrieves relevant evidence from the literature and other sources and generates grounded outputs with citations, which are then surfaced to scientists in the user interface of this safety-focused tool.

\section{Conclusion}
Research Assistant was designed around three practical priorities: data accuracy, response speed, and low operating cost. It occupies a different space from more open-ended autonomous research systems, such as Google CoScientist \cite{cosci_paper} or Robin \cite{robin}, and is instead aimed at helping scientists and clinicians quickly access grounded biomedical information in day-to-day R\&D work. Research Assistant can also serve as a grounded biomedical data endpoint for larger agentic systems through its MCP and REST API endpoints. Important challenges remain. These include hallucinations and limited sensitivity to biological nuance, such as distinctions between closely related gene paralogs. As methods for agentic AI and the engineering of reliable research systems continue to improve, we expect these limitations to become more manageable. Given the scale and complexity of biomedical knowledge, such integrative systems are likely to play an increasingly important role in helping researchers identify relevant signals within large and fragmented information spaces.

\bibliographystyle{plain}
\bibliography{references}

\end{document}